\documentclass[a4paper,twoside]{article}
\usepackage[utf8]{inputenc}
\usepackage{algorithm2e}
\usepackage{times}
\usepackage{epsfig}
\usepackage{graphicx}
\usepackage{amsmath}
\usepackage{amssymb}
\usepackage{multirow}
\usepackage{hyphenat}
\usepackage{rotating}
\usepackage{afterpage}
\usepackage{color}
\usepackage{tabu}
\usepackage{hyphenat}
\usepackage{color, colortbl}
\usepackage{multirow}
\usepackage{booktabs}
\usepackage{caption}
\usepackage{epsfig}
\usepackage{calc}
\usepackage{amssymb}
\usepackage{amstext}
\usepackage{amsmath}
\usepackage{amsthm}
\usepackage{multicol}
\usepackage{pslatex}
\usepackage{apalike}
\usepackage{SCITEPRESS}     

\begin{document}


\title{AuxNet: Auxiliary tasks enhanced \\
Semantic Segmentation for Automated Driving}


 \author{\authorname{Sumanth Chennupati$^{1,3}$, Ganesh Sistu\sup{2}, Senthil Yogamani\sup{2} and  Samir Rawashdeh\sup{3}}
  \affiliation{\sup{1}Valeo Troy, United States}
  \affiliation{\sup{2}Valeo Vision Systems, Ireland} \affiliation{\sup{3}University of Michigan-Dearborn}
  \email{\{sumanth.chennupati, ganesh.sistu,senthil.yogamani\}@valeo.com, srawa@umich.edu}  }


\abstract{
Decision making in automated driving is highly specific to the environment and thus semantic segmentation plays a key role in recognizing the objects in the environment around the car. Pixel level classification once considered a challenging task which is now becoming mature to be productized in a car. However, semantic annotation is time consuming and quite expensive. Synthetic datasets with domain adaptation techniques have been used to alleviate the lack of large annotated datasets. In this work, we explore an alternate approach of leveraging the annotations of other tasks to improve semantic segmentation. Recently, multi-task learning became a popular paradigm in automated driving which demonstrates joint learning of multiple tasks improves overall performance of each tasks. Motivated by this, we use auxiliary tasks like depth estimation to improve the performance of semantic segmentation task. We propose adaptive task loss weighting techniques to address scale issues in multi-task loss functions which become more crucial in auxiliary tasks. We experimented on automotive datasets including SYNTHIA and KITTI and obtained 3\% and 5\% improvement in accuracy respectively.   
}

\onecolumn \maketitle \normalsize \vfill


\section{Introduction}
Semantic image segmentation has witnessed tremendous progress recently with deep learning. It provides dense pixel-wise labeling of the image which leads to scene understanding. Automated driving is one of the main application areas where it is commonly used. The level of maturity in this domain has rapidly grown recently and the computational power of embedded systems have increased as well to enable commercial deployment. Currently, the main challenge is the cost of constructing large datasets as pixel-wise annotation is very labor intensive. It is also difficult to perform corner case mining as it is a unified model to detect all the objects in the scene. Thus there is a lot of research to reduce the sample complexity of segmentation networks by incorporating domain knowledge and other cues where-ever possible. One way to overcome this is via using synthetic datasets and domain adaptation techniques \cite{Sankaranarayanan2018LearningFS}, another way is to use multiple clues or annotations to learn efficient representations for the task with limited or expensive annotations \cite{liebel2018auxiliary}. 

\begin{figure}[!t]
\centering
\includegraphics[width=0.48\textwidth,height=8em]{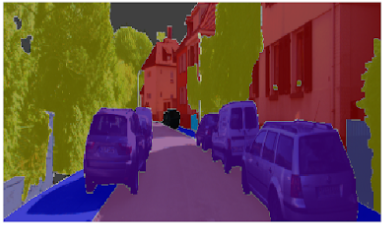}
\caption{Semantic Segmentation of an automotive scene}
\label{fig:segnet}
\end{figure} 

In this work, we explore the usage of auxiliary task learning to improve the accuracy of semantic segmentation. We demonstrate the improvements in semantic segmentation by inducing depth cues via auxiliary learning of depth estimation. The closest related work is \cite{liebel2018auxiliary} where auxiliary task was used to improve semantic segmentation task using GTA game engine. Our work demonstrates it for real and synthetic datasets using novel loss functions. The contributions of this work include: 

\begin{enumerate}
    \item Construction of auxiliary task learning architecture for semantic segmentation.
    \item Novel loss function weighting strategy for one main task and one auxiliary task.
    \item Experimentation on two automotive datasets namely KITTI and SYNTHIA.
\end{enumerate}

The rest of the paper is organized as follows: Section \ref{sec:back} reviews the background in segmentation in automated driving and learning using auxiliary tasks. Section \ref{sec:method} details the construction of auxiliary task architecture and proposed loss function weighting strategies. Section \ref{sec:exps} discusses the experimental results in KITTI and SYNTHIA. Finally, section \ref{sec:conc} provides concluding remarks.

\section{Background} \label{sec:back}
\subsection{Semantic Segmentation} 

A detailed survey of semantic segmentation for automated driving is presented in \cite{8317714}. We briefly summarize the relevant parts focused on CNN based methods. FCN \cite{long2015fully} is the first CNN based end to end trained pixel level segmentation network. Segnet \cite{Badrinarayanan2017SegNetAD} introduced encoder decoder style semantic segmentation. U-net \cite{cciccek20163d} is also an encoder decoder network with dense skip connections between the them. While these papers focus on architectures, Deeplab \cite{7913730} and EffNet \cite{8451339} focused on efficient convolutional layers by using dilated and separable convolutions.

Annotation for semantic segmentation is a tedious and expensive process. An average experienced annotator takes anywhere around 10 to 20 minutes for one image and it takes 3 iterations for correct annotations, this process limit the availability of large scale accurately annotated datasets. Popular semantic segmentation automotive datasets like CamVid \cite{BrostowFC:PRL2008}, Cityscapes \cite{Cordts2016Cityscapes}, KITTI \cite{Kitti} are relatively smaller when compared to classification datasets like ImageNet \cite{imagenet_cvpr09}. Synthetic datasets like Synthia \cite{ros2016synthia}, Virtual KITTI \cite{Gaidon:Virtual:CVPR2016}, Synscapes \cite{Wrenninge2018SynscapesAP} offer larger annotated synthetic data for semantic segmentation. Efforts like Berkley Deep Drive \cite{Xu2017EndtoEndLO}, Mapillary Vistas \cite{neuhold2017mapillary} and Toronto City \cite{Wang2017TorontoCityST} have provided larger datasets to facilitate training a deep learning model for segmentation.   

\subsection{Multi-Task Learning}

Multi-task learning \cite{8100062}, \cite{Chen_2018}, \cite{neven2017fast} has been gaining significant popularity over the past few years as it has proven to be very efficient for embedded deployment. Multiple tasks like object detection, semantic segmentation, depth estimation etc can be solved simultaneously using a single model. A typical multi-task learning framework consists of a shared encoder coupled with multiple task dependent decoders. An encoder extracts feature vectors from an input image after series of convolution and poling operations. These feature vectors are then processed by individual decoders to solve different problems. \cite{Teichmann_2018} is an example where three task specific decoders were used for scene classification, object detection and road segmentation of an automotive scene. The main advantages of multi-task learning are improved computational efficiency, regularization and scalability. \cite{ruder2017overview} discusses other benefits and applications of multi-task learning in various domains.      

\begin{figure*}[!t]
\centering
\includegraphics[width=0.75\textwidth,height=20em]{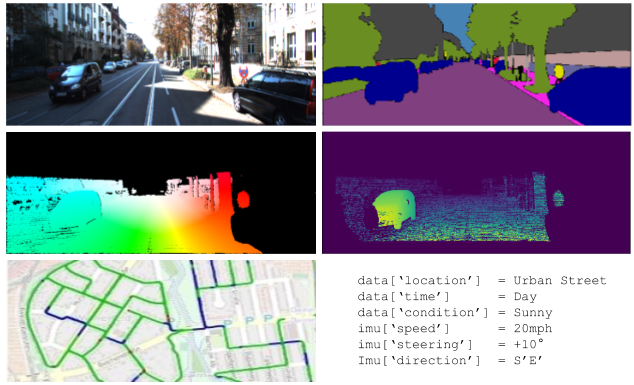}
\caption{Illustration of several Auxiliary Visual Perception tasks in an Automated driving dataset KITTI. First Row shows RGB and Semantic Segmentation, Second Row shows Dense Optical Flow and Depth, Third row shows Visual SLAM and meta-data for steering angle, location and condition}
\label{fig:auxtasks}
\end{figure*}

\subsection{Auxiliary Task Learning}
Learning a side or auxiliary task jointly during training phase to enhance main task's performance is usually referred to auxiliary learning. This is similar to multi-task learning except the auxiliary task is nonoperational during inference. This auxiliary task is usually selected to have much larger annotated data so that it acts a regularizer for main task. In \cite{liebel2018auxiliary} semantic segmentation is performed using auxiliary tasks like time, weather, etc. In \cite{Toshniwal2017MultitaskLW}, end2end speech recognition training uses auxiliary task phoneme recognition for initial stages. \cite{Parthasarathy2018LadderNF} uses unsupervised aux tasks for audio based emotion recognition. It is often believed that auxiliary tasks can be used to focus attention on a specific parts of the input. Predictions of road characteristics like markings as an auxiliary task in \cite{caruana1997multitask} to improve main task for steering prediction is one instance of such behaviour. 

Figure \ref{fig:auxtasks} illustrates auxiliary tasks in a popular automated driving dataset KITTI. It contains various dense output tasks like Dense optical flow, depth estimation and visual SLAM. It also contains meta-data like steering angle, location and external condition. These meta-data comes for free without any annotation task. Depth could be obtained for free by making use of Velodyne depth map, \cite{kumar2018monocular} demonstrate training using sparse Velodyne supervsion.

{\color{red}






}

\subsection{Multi-Task Loss} \label{multiloss}
Modelling a multi-task loss function is a critical step in multi-task training. An ideal loss function should enable learning of multiple tasks with equal importance irrespective of loss magnitude, task complexity etc.  Manual tuning of task weights in a loss function is a tedious process and it is prone to errors. Most of the work in multi-task learning uses a linear combination of multiple task losses which is not effective.   \cite{kendall2017multi} propose an approach to learn the optimal weights adaptively based on uncertainty of prediction. The log likelihood of the proposed joint probabilistic model shows that the task weights are inversely proportional to the uncertainty. Minimization of total loss w.r.t task uncertainties and losses converges to an optimal loss weights distribution.  This enables independent tasks to learn at a similar rate allowing each to influence on training. However, these task weights are adjusted at the beginning of the training and are not adapted during the learning. GradNorm \cite{Chen2018GradNormGN} proposes an adaptive task weighing approach by normalizing gradients from each task. They also consider the rate of change of loss to adjust task weights. \cite{liu2018endtoend} adds a moving average of task weights obtained by method similar to GradNorm. \cite{guo2018dynamic} on other hand proposes dynamic weight adjustments based on task difficulty. As the difficulty of learning changes over training time, the task weights are updated allowing the model to prioritize difficult tasks.      Modelling multi-task loss as a multi-objective function was proposed in \cite{Zhang2010ACF}, \cite{sener2018multitask} and \cite{Dsidri2009MultipleGradientDA}.  A reinforcement learning approach was used in \cite{liu2018exploration} to minimize the total loss while changing the loss weights. 
 
 \begin{figure*}[!t]
\centering
\includegraphics[width=0.95\textwidth,height=17em]{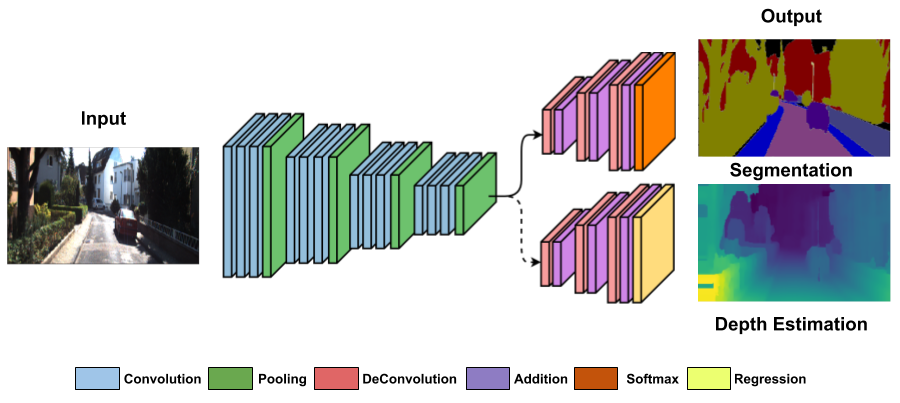}
\caption{AuxNet: Auxiliary Learning network with Segmentation as main task and Depth Estimation as auxiliary task.}
\label{fig:auxnet}
\end{figure*}

\section{Methods} \label{sec:method}
Semantic segmentation and depth estimation have common feature representations. Joint learning of these tasks have shown significant performance gains in \cite{liu2018endtoend}, \cite{Eigen_2015}, \cite{Mousavian_2016}, \cite{Jafari_2017} and \cite{Gurram_2018}. Learning underlying representations between these tasks help the multi-task network alleviate the confusion in predicting semantic boundaries or depth estimation. Inspired by these papers, we propose a multi-task network with semantic segmentation as main task and depth estimation as an auxiliary task. As accuracy of the auxiliary task is not important, weighting its loss function appropriately is important. We also discuss in detail the proposed auxiliary learning network and how we overcame the multi-task loss function challenges discussed in section \ref{multiloss}.  

\subsection{Architecture Design} \label{arch}
The proposed network takes input RGB image and outputs semantic and depth maps together. Figure \ref{fig:auxnet} shows two task specific decoders coupled to a shared encoder to perform semantic segmentation and depth estimation. The shared encoder is built using ResNet-50 \cite{7780459} by removing the fully connected layers from the end. The encoded feature vectors are now passed to two parallel stages for independent task decoding. Semantic segmentation decoder is constructed similar to FCN8 \cite{long2015fully} architecture with transposed convolutions, up sampling and skip connections. The final output is made up of a softmax layer to output probabilistic scores for each semantic class. Depth estimation decoder is also constructed similar to segmentation decoder except the final output is replaced with a regression layer to estimate scalar depth.

\subsection{Loss Function } \label{lossfun}
In general, a multi-task loss function is expressed as weighted combination of multiple task losses where $\mathcal{L}_{i}$ is loss and $\lambda_{i}$ is associated weight for task $i$. 
\begin{equation}\label{eq:1}
\mathcal{L}_{Total} = \displaystyle\sum_{i=1}^{t} \lambda_{i}\mathcal{L}_{i}  
\end{equation}

For the proposed 2-task architecture we express loss as:
\begin{equation}\label{eq:2}
\mathcal{L}_{Total} =  \lambda_{Seg}\mathcal{L}_{Seg} + \lambda_{Depth}\mathcal{L}_{Depth}  
\end{equation}

$\mathcal{L}_{Seg}$ is semantic segmentation loss expressed as an average of pixel wise cross-entropy for each predicted label and ground truth label. $\mathcal{L}_{Depth}$ is depth estimation loss expressed as mean absolute error between estimated depth and true depth for all pixels. To overcome the significant scale difference between semantic segmentation and depth estimation losses, we perform task weight balancing as proposed in Algorithm \ref{algo:1}.  Expressing multi-task loss function as product of task losses, forces each task to optimize so that the total loss reaches a minimal value. This ensures no task is left in a stale mode while other tasks are making progress. By making an update after every batch in an epoch, we dynamically change the loss weights. We also add a moving average to the loss weights to smoothen the rapid changes in loss values at the end of every batch. 

\begin{algorithm}
\caption{Proposed Weight Balancing for 2-task semantic segmentation and depth estimation}
\label{algo:1}
\SetAlgoLined
 \For{$epoch\gets1$ \KwTo $n$}{
    \For{$batch\gets1$ \KwTo $s$}{
\begin{math}    
    \lambda_{Seg} \hspace{.85em}   = \mathcal{L}_{Depth} \newline
    \lambda_{Depth} = \mathcal{L}_{Seg} \newline
    \mathcal{L}_{Total} \hspace{.15em} = \mathcal{L}_{Depth}\mathcal{L}_{Seg} + \mathcal{L}_{Seg}\mathcal{L}_{Depth} \newline
    \mathcal{L}_{Total}\hspace{.15em} = 2\times \mathcal{L}_{Seg}\mathcal{L}_{Depth}
\end{math}
}}
\end{algorithm}

\begin{figure*}[ht]
\centering
\includegraphics[width=.9\textwidth]{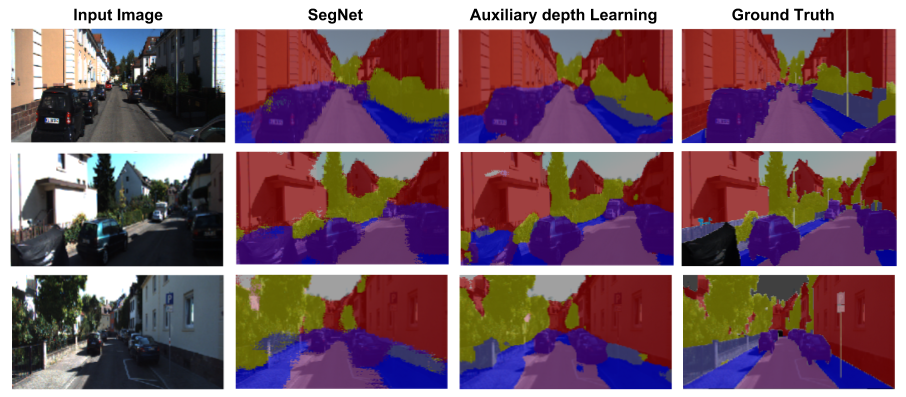}
\label{fig:Kitti}
\centering
\includegraphics[width=.9\textwidth]{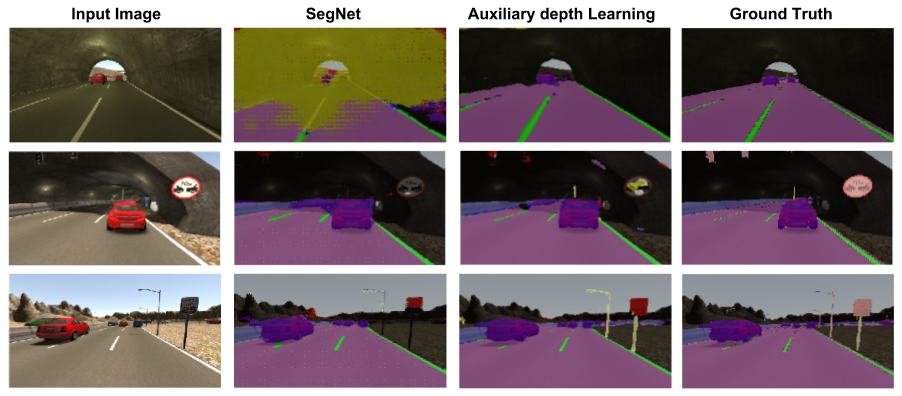}
\label{fig:Synth}
\caption{Results on KITTI and  SYNTHIA datasets}
\label{fig:Results}
\end{figure*}

In Algorithm \ref{algo:2}, we propose focused task weight balancing to prioritize the main task's loss in auxiliary learning networks. We introduce an additional term to increase the weight of main task. This term could be a fixed value to scale up main task weight or the magnitude of task loss.      

\begin{algorithm}
\caption{Proposed Focused Task Weight Balancing for Auxiliary Learning.}
\label{algo:2}
\SetAlgoLined
 \For{$epoch\gets1$ \KwTo $n$}{
    \For{$batch\gets1$ \KwTo $s$}{
\begin{math}     
    \lambda_{Seg}\hspace{.85em} = \mathcal{L}_{Seg} \times \mathcal{L}_{Depth} \newline
    \lambda_{Depth} = \mathcal{L}_{Seg} \newline
    \mathcal{L}_{Total}\hspace{.15em} = \mathcal{L}_{Seg}^{2}\mathcal{L}_{Depth} + \mathcal{L}_{Seg}\mathcal{L}_{Depth} \newline
    \mathcal{L}_{Total}\hspace{.15em} = (\mathcal{L}_{Seg}+1)\times\mathcal{L}_{Seg}\mathcal{L}_{Depth}
\end{math}
}}
\end{algorithm}

\section{Results and Discussion} 

\label{sec:exps}
In this section, we present details about the experimental setup used and  discuss the observations on the results obtained.
\subsection{Experimental Setup}
We implemented the auxiliary learning network as discussed in section \ref{arch} to perform semantic segmentation and depth estimation. We chose ResNet-50 as the shared encoder which is pre-trained on ImageNet. We used segmentation and depth estimation decoders with random weight initialization. We performed all our experiments on KITTI \cite{Kitti} semantic segmentation and SYNTHIA \cite{ros2016synthia} datasets. These datasets contain RGB image data, ground truth semantic labels and  depth data represented as disparity values in 16-bit png format.  We re-sized all the input images to a size 224x384. 

\begin{table*}[!ht]
\centering
\caption{Comparison Study : SegNet vs different auxiliary networks.}
\label{table:auxnet}
\resizebox{\textwidth}{!}{%
\begin{tabular}{lcccccccccc}
\hline
\hline
\multicolumn{11}{c}{\textbf{KITTI}} \\ \hline
\textbf{Model} & \textbf{Sky} & \textbf{Building} & \textbf{Road} & \textbf{Sidewalk} & \textbf{Fence} & \textbf{Vegetation} & \textbf{Pole} & \textbf{Car} & \textbf{Lane} & \textbf{IoU} 
\\ \hline
SegNet & 46.79 & 87.32 & 89.05 & 60.69 & \textbf{22.96} & 85.99 & - & 74.04 & - & 74.52
\\
$AuxNet_{400}$ & 49.11 & 88.55 & 93.17 & 69.65 & 22.93 & 87.12 & - & 74.63 & - & 78.32 
\\
$AuxNet_{1000}$ & 49.17 & 89.81 & 90.77 & 64.16 & 14.77 & 86.52 & - & 71.40 & - & 76.58 
\\ 
$AuxNet_{\rm TWB}$ & \textbf{49.73} & \textbf{91.10} & 92.30 & 70.55 & 18.64 & 86.01 & - & 77.32 & - & 78.64
\\ 
$AuxNet_{\rm FTWB}$ & 48.43 & 89.50 & \textbf{92.71} & \textbf{71.58} & 15.37 &\textbf{ 88.31} & - & \textbf{79.55} & - & \textbf{79.24} 
\\ 
\hline
\hline
\multicolumn{11}{c}{\textbf{SYNTHIA}} \\ \hline
\textbf{Model} & \textbf{Sky} & \textbf{Building} & \textbf{Road} & \textbf{Sidewalk} & \textbf{Fence} & \textbf{Vegetation} & \textbf{Pole} & \textbf{Car} & \textbf{Lane} & \textbf{IoU} \\ \hline
SegNet & 95.41 & 58.18 & 93.46 & 09.82 & 76.04 & 80.95 & 08.79 & 85.73 & 90.28 & 89.70  \\
$AuxNet_{400}$ & 95.12 & \textbf{69.82} & 92.95 & 21.38 & 77.61 & 84.23& 51.31 &\textbf{90.42 }& 91.20 & 91.44 \\
$AuxNet_{1000}$ & 95.41 & 59.57 & \textbf{96.83} & 28.65 & \textbf{81.23} & 82.48 & \textbf{56.43} & 88.93 & 94.19 & \textbf{92.60}\\
$AuxNet_{\rm TWB}$ & 94.88 & 66.41 & 94.81 & \textbf{31.24} & 77.01 & \textbf{86.04} & 21.83 & 90.16 & \textbf{94.47 }& 91.67 
\\ 
$AuxNet_{\rm FTWB}$ & \textbf{95.82} & 56.19 & 96.68 & 21.09 & 81.19 & 83.26 & 55.86 & 89.01 & 92.11 & 92.05 
\\
\hline
\hline
\end{tabular}
}
\end{table*}

The loss function is expressed as detailed in section \ref{lossfun}. Categorical cross-entropy was used to compute semantic segmentation loss and mean absolute error is used to compute depth estimation loss. We implemented four different auxiliary learning networks by changing the expression of loss function. AuxNet$_{400}$ and AuxNet$_{1000}$ weighs segmentation loss 400 and 1000 times compared to depth estimation loss. AuxNet$_{\rm TWB}$ and AuxNet$_{\rm FTWB}$ are built based on Algorithms \ref{algo:1} and \ref{algo:2} respectively. These networks are trained with ADAM \cite{kingma2014adam} optimizer for 200 epochs. The best model for each network was saved by monitoring the validation loss of semantic segmentation task. Mean IoU and categorical IoU were used for comparing the performance.  

\subsection{Results and Discussion}

In Table \ref{table:auxnet}, we compare the proposed auxiliary learning networks (AuxNet) against a simple semantic segmentation network (SegNet) constructed using an encoder decoder combination. The main difference between these two networks is the additional depth estimation decoder. It is observed that auxiliary networks perform better than the baseline semantic segmentation. It is evident that incorporating depth information improves the performance of segmentation task. It is also observed that depth dependent categories like sky, sidewalk, pole and car have shown better improvements than other categories due to availability of depth cues.  

\begin{table}[!ht] 
\centering
\caption{Comparison between SegNet, FuseNet and AuxNet in terms of performance and parameters.}
\label{table:fusenet}
\begin{tabular}{lcc}
\hline
\hline
\multicolumn{3}{c}{\textbf{KITTI}} 
\\ \hline
\textbf{Model} & \textbf{IoU} & \textbf{Params} 
\\ \hline
SegNet & 74.52 & 23,672,264 \\
FuseNet & \textbf{80.99} & 47,259,976 \\
AuxNet & 79.24 & 23,676,142 \\
\hline
\hline
\multicolumn{3}{c}{\textbf{SYNTHIA}} \\
\hline
\textbf{Model} & \textbf{IoU} & \textbf{Params} \\ \hline
SegNet & 89.70 & 23,683,054 \\
FuseNet & 92.52 & 47,270,766 \\
AuxNet & \textbf{92.60} & 23,686,932\\
\hline
\hline
\end{tabular}
\end{table}

 We compare the performances of SegNet, AuxNet with FuseNet in Table \ref{table:fusenet}. FuseNet is another semantic segmentation network (FuseNet) that takes RGB images and depth map as input. It is constructed in a similar manner to the work in \cite{hazirbas2016fusenet}. We compare the mean IoU of each network and the number of parameters needed to construct the network. AuxNet required negligible increase in parameters while FuseNet almost needed twice the number of parameters compared to SegNet. It is observed AuxNet can be chosen as a suitable low cost replacement to FuseNet as the needed depth information is learned by shared encoder.       

\section{Conclusion} \label{sec:conc}
Semantic segmentation is a critical task to enable fully automated driving. It is also a complex task and requires large amounts of annotated data which is expensive. Large annotated datasets is currently the bottleneck for achieving high accuracy for deployment. In this work, we look into an alternate mechanism of using auxiliary tasks to alleviate the lack of large datasets. We discuss how there are many auxiliary tasks in automated driving which can be used to improve accuracy. We implement a prototype and use depth estimation as an auxiliary task and show 5\% improvement on KITTI and 3\% improvement on SYNTHIA datasets. We also experimented with various weight balancing strategies which is a crucial problem to solve for enabling more auxiliary tasks. In future work, we plan to augment more auxiliary tasks.




\bibliographystyle{apalike}
{\small
\bibliography{acmart}
}

\end{document}